\documentclass[runningheads]{llncs}

\usepackage[mobile]{eccv}

\usepackage{eccvabbrv}

\usepackage{graphicx}
\usepackage{bm} 
\usepackage[b]{esvect}  
\newcommand{\svec}[1]{\vv{#1}} 
\usepackage{booktabs}
\usepackage{ulem}
\usepackage[accsupp]{axessibility}  

\usepackage[pagebackref,breaklinks,colorlinks,citecolor=eccvblue]{hyperref}

\usepackage{orcidlink}
\usepackage{marvosym}

\usepackage{multirow}
\usepackage{pifont}
\newcommand{\customfootnotetext}[1]{%
  \begingroup
    \renewcommand{\thefootnote}{}
    \footnotetext{#1}%
  \endgroup
}

\begin{document}


\title{UniStitch: Unifying Semantic and Geometric Features for Image Stitching} 

\titlerunning{Abbreviated paper title}

\author{Yuan Mei\inst{1,2} \and
Lang Nie\textsuperscript{1,\ding{81}} \and
Kang Liao\inst{3} \and
Yunqiu Xu\inst{4} \and \\
Chunyu Lin\inst{5} \and
Bin Xiao\inst{1}
}

\authorrunning{Y.~Mei et al.}

\institute{\textsuperscript{1}CQUPT, \textsuperscript{2}PolyU, \textsuperscript{3}NTU, \textsuperscript{4}NUS, \textsuperscript{5}BJTU\\
\email{yyuan.mei@connect.polyu.hk, nielang@cqupt.edu.cn} \\
Project Page: \url{https://mmelodyy.github.io/projects/unistitch}
}

\customfootnotetext{Work done during the internship at CQUPT. \textsuperscript{{\ding{81}}}Project lead.}

\maketitle

\begin{figure}[!htbp]
    \centering      
    \vspace{-0.2cm}
    \begin{minipage}{0.98\linewidth}
        \centering
        \includegraphics[width=\linewidth]{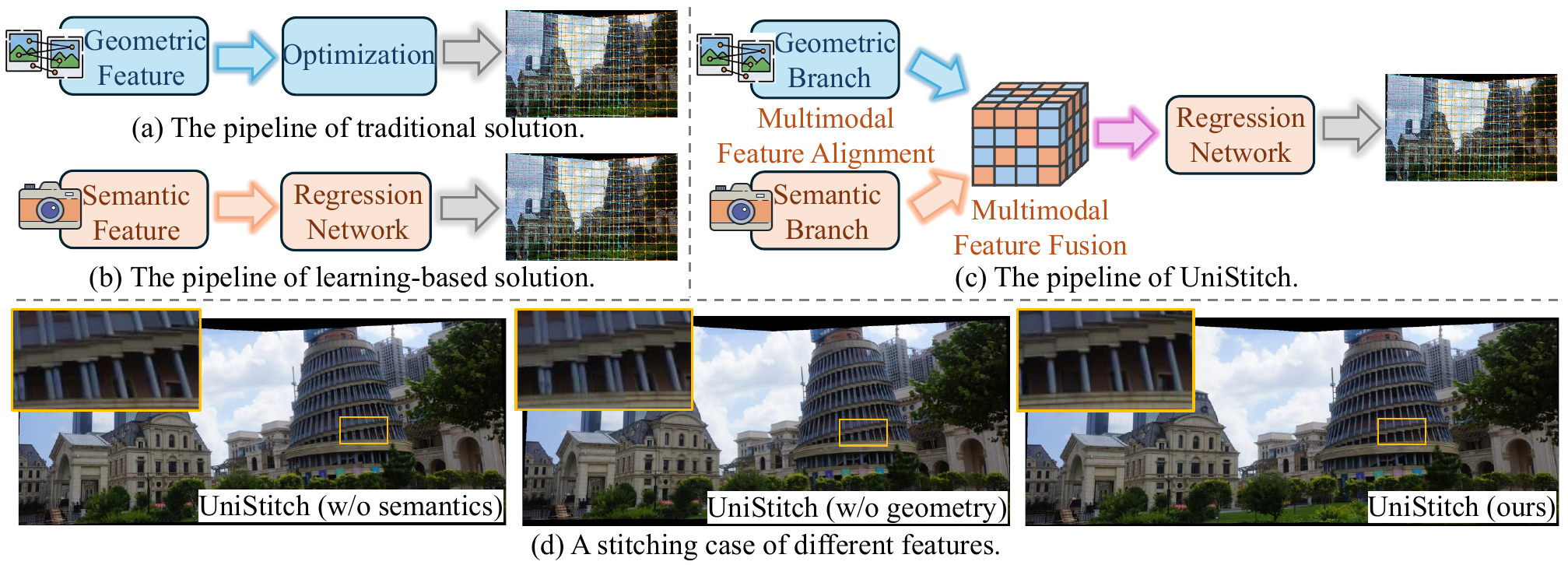}
        \caption{Existing image stitching solutions vs. UniStitch (ours). The proposed UniStitch integrates the traditional geometric feature and learning-based semantic feature into a unified representation. It effectively eliminates artifacts that persist when either feature modality is used alone, delivering a clear improvement.}
        \vspace{-0.7cm}
        \label{fig:teaser}
    \end{minipage}
\end{figure}

\begin{abstract}
Traditional image stitching methods estimate warps from hand-crafted geometric features, whereas recent learning-based solutions leverage semantic features from neural networks instead. These two lines of research have largely diverged along separate evolution, with virtually no meaningful convergence to date. In this paper, we take a pioneering step to bridge this gap by unifying semantic and geometric features with \textbf{UniStitch}, a \underline{uni}fied image \underline{stitch}ing framework from multimodal features. To align discrete geometric features (\textit{i.e.}, keypoint) with continuous semantic feature maps, we present a Neural Point Transformer (NPT) module, which transforms unordered, sparse 1D geometric keypoints into ordered, dense 2D semantic maps. Then, to integrate the advantages of both representations, an Adaptive Mixture of Experts (AMoE) module is designed to fuse geometric and semantic representations. It dynamically shifts focus toward more reliable features during the fusion process, allowing the model to handle complex scenes, especially when either modality might be compromised. The fused representation can be adopted into common deep stitching pipelines, delivering significant performance gains over any single feature. Experiments show that UniStitch outperforms existing state-of-the-art methods with a large margin, paving the way for a unified paradigm between traditional and learning-based image stitching.

\end{abstract}

\section{Introduction}
Image stitching is a fundamental technique in computer vision, enabling the creation of seamless panoramic images from multiple overlapping views. It plays a critical role in a wide range of applications, including virtual reality, autonomous driving, medical imaging, and computational photography.

Existing solutions are generally divided into two categories. Traditional approaches~\cite{brown2007automatic,zaragoza2013projective} (Fig. \ref{fig:teaser}a) leverage hand-crafted geometric features (\textit{e.g.}, SIFT~\cite{lowe2004distinctive}) to establish correspondences and then optimize the desired warps. They perform reliably in structured, texture-rich scenes, yet struggle in low-texture or repetitive environments where feature detection becomes unreliable. In contrast, learning-based methods~\cite{nie2021unsupervised,nie2023parallax} (Fig. \ref{fig:teaser}b) adopt deep neural networks to extract semantic features, which focus on high-level content understanding of the scene. While this enhances robustness in challenging visual conditions, it also means their learned representations are oriented towards scene understanding rather than capturing geometric structures. Consequently, even in well-structured scenes, where traditional methods perform well, current deep stitching techniques do not consistently demonstrate superiority. This gap naturally raises a critical question: \textit{Where does the future of image stitching lie—in geometric or semantic features?}

In this paper, we answer this by introducing \textbf{UniStitch}, a unified image stitching framework from both semantic and geometric features.
As shown in Fig. \ref{fig:teaser}c, UniStitch consists of three stages: multi-modality feature alignment, multi-modality feature fusion, and global-to-local warp. Here, we take the most commonly used geometric feature, \textit{i.e.}, keypoint, as the representative and demonstrate how to combine the two totally different modality features.

\textbf{The first stage} is composed of a semantic branch and a geometric branch, which extract multimodal features from images and keypoints, respectively. To align the discrete keypoints with continuous semantic feature maps, we propose a Neural Point Transformer (NPT) module to encode 1D points into 2D geometric maps. It first transforms shallow keypoints into high-dimensional point features, and then projects them into a structured latent space (\textit{i.e.}, a mesh-like representation) by explicitly reorganizing their spatial relationships. 
Then, \textbf{the second stage} is designed to fuse geometric and semantic features to construct a robust joint representation. To this end, we design an Adaptive Mixture of Experts (AMoE) module that adaptively captures the heterogeneity of multi-modal features and integrates their complementary strengths. Simultaneously, we implement a Latent-space Modality Robustifier (MR) strategy to equip the model with cross-scene robustness, particularly when a certain modality becomes unreliable.
\textbf{The third stage} predicts the parametric warp following a global-to-local paradigm. Departing from standard practices~\cite{nie2025stabstitch++}, we introduce a Free-Form Deformation (FFD) module designed to enhance the efficiency of Thin-Plate Spline (TPS) transformations for high-resolution images. This approach significantly reduces VRAM overhead and accelerates inference while preserving precise spatial alignment.

Experiments demonstrate UniStitch achieves SoTA performance across diverse datasets by integrating the strengths of different modality features. Furthermore, it can be combined with different geometric features and make sense, contributing to a pioneering and meaningful step for unified image stitching.

\section{Related Work}
Here, we review existing image stitching solutions from the perspectives of geometric and semantic features, respectively.

\subsection{Traditional Stitching using Geometric Features}
Traditional image stitching pipelines typically begin by detecting and matching local geometric features—most classically, keypoints such as SIFT~\cite{lowe2004distinctive}. The seminal Autostitch work~\cite{brown2007automatic} demonstrated that a global homography could robustly align images based on such point correspondences. To handle more complex parallax, researchers subsequently extended the warp model beyond global homography to various elastic representations, including mesh-based warps~\cite{zaragoza2013projective,zhang2020content}, thin-plate splines (TPS)~\cite{li2017parallax}, superpixel-based deformations~\cite{lee2020warping}, and local triangular facets~\cite{li2019local}. Alongside this, the feature basis itself has expanded from keypoints to include line segments~\cite{li2015dual}, edges~\cite{du2022geometric}, and other structural cues. This evolution not only enhances matching robustness in structured scenes but also actively contributes to preserving geometric shape consistency. These include smoothly transitioning warps from projective to similarity transformation~\cite{chang2014shape,lin2015adaptive}, enforcing line consistency across the warping process~\cite{li2015dual,liao2019single,jia2021leveraging}, and imposing a global similarity prior~\cite{chen2016natural}. 
More recently, some researchers have begun to incorporate higher-level semantic information, notably semantic segmentation, to guide the stitching process. For instance, OBJ-GSP\cite{cai2025object} adopts segmentation maps to extract the contours of any objects and then preserves their structures, while MHW\cite{liao2025parallax} and PRC\cite{li2021image} reorganize geometric correspondences based on segmentation consistency to improve alignment in parallax scenes. This, however, remains a preliminary fusion of geometry and semantics. The semantic prior is used merely as a static guide within the optimization, rather than being jointly encoded and interactively learned with geometric features at the representational level, leaving the two modalities fundamentally isolated.

\subsection{Learning-based Stitching using Semantic Features}
Driven by the success of deep learning, recent stitching methods have shifted toward end-to-end trainable pipelines. These approaches typically feed image pairs or sequences directly into neural networks, which are trained to predict parametric warps, such as homography~\cite{nie2020view,nie2022learning,jiang2022towards}, multiple homography~\cite{song2021end,nie2021depth}, TPS~\cite{nie2023parallax,kim2024learning,liao2023recrecnet,nie2025stabstitch++,zhang2024recstitchnet,liao2025mowa}, or dense optical flow~\cite{kweon2023pixel,jia2023learning,jin2025pixelstitch}. By establishing dense, pixel-wise correspondence objectives, these networks learn to extract and align semantic features, gaining notable robustness in challenging scenarios where traditional geometric features are scarce, such as adverse conditions, low-texture, low-light, or repetitive environments~\cite{zhang2025image,jiang2022towards,jiang2024towards,liao2025thinking}. However, this very strength also reveals a limitation: the current research is heavily skewed toward addressing these “hard cases”, often at the expense of performance in well-structured, geometric-rich scenes where traditional solutions demonstrate reliable performance. The learned representations, focused on high-level content understanding, exhibit a fundamental divergence from explicit geometric structure. Recently, RopStitch~\cite{nie2026robust} further digs into semantic cues by introducing extra content understanding prior, yet still operates predominantly within the semantic feature space. They fall short of explicitly integrating the well-established geometric features inherent to the scene, underscoring a persisting gap: the absence of a unified framework that merges both semantic and geometric features.

\section{UniStitch}
UniStitch follows a hierarchical three-stage pipeline consisting of multimodal feature alignment, multimodal feature fusion, and global-to-local warp. 

\subsection{Overview}
\begin{figure}[htpb!] 
	\centering  
	\includegraphics[width=\linewidth]{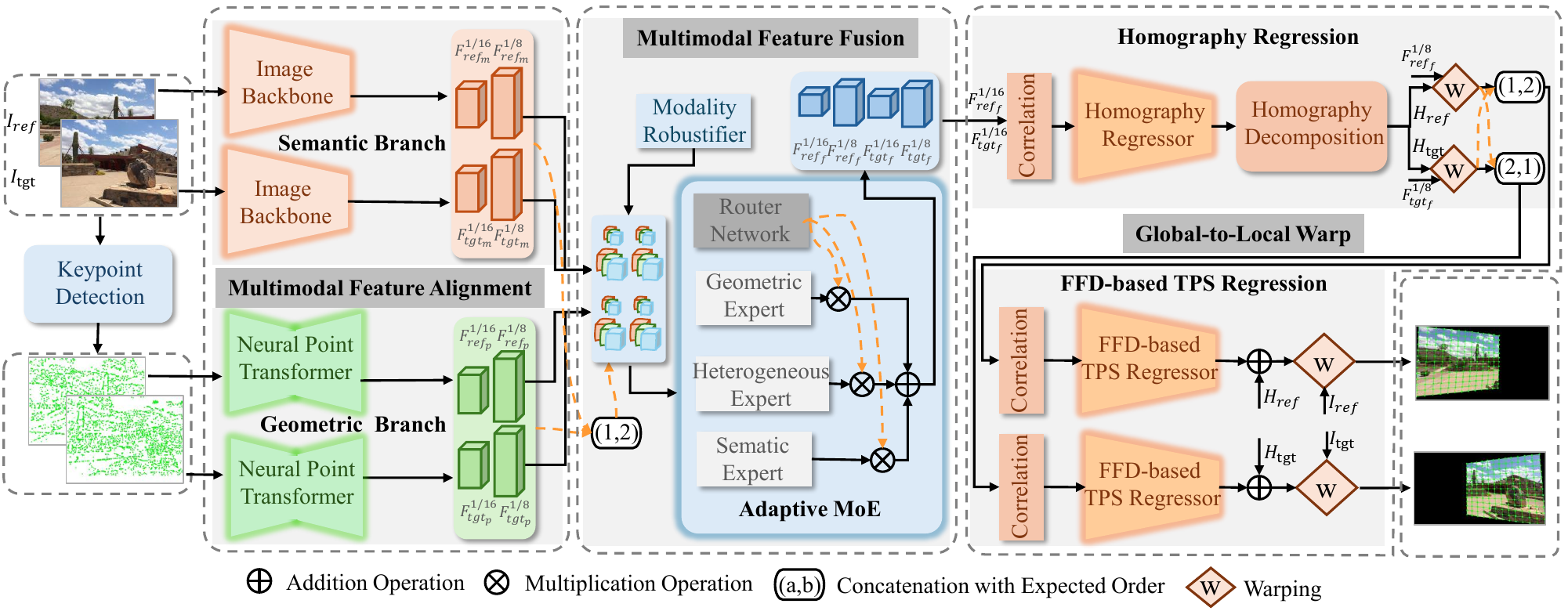} 
	\caption{The overall structure of UniStitch. It takes original images and their keypoints as input, and follows a three-stage pipeline: multimodal feature alignment, multimodal feature fusion, and global-to-local warp.}
	\label{fig:network}
\end{figure}

As illustrated in Fig.~\ref{fig:network}, we first extract geometric keypoints and descriptors ($P_\text{ref}$, $P_\text{tgt}$) from a pair of reference and target images ($I_\text{ref}$, $I_\text{tgt}$). The multimodal feature alignment stage performs intra-modal feature extraction and inter-modal feature alignment. It is composed of two branches, in which the semantic branch extracts high-level semantic maps (denoted as $F_\text{ref,s}$, $F_\text{tgt,s}$) from the image domain, while the geometric branch leverages a neural point transformer module to convert sparse, unordered keypoints into dense, structured geometric maps (denoted as $F_\text{ref,g}, F_\text{tgt,g}$). In the multimodal feature fusion stage, we design an adaptive mixture of experts module with a latent-space modality robustifier that intelligently integrates multimodal cues by weighting their reliability, yielding robust fused features (denoted as  $F_\text{ref,f}$, $F_\text{tgt,f}$). Finally, the global-to-local warp stage adopts a similar warping strategy to StabStitch++~\cite{nie2025stabstitch++}, where we incorporate free-form deformation to optimize the vanilla TPS transformation. This crucial modification effectively reduces VRAM consumption and accelerates inference speed while maintaining accurate warping capability.

\subsection{Multimodal Feature Alignment}
Aligning multimodal features with a unified representation is the first step of UniStitch.
To obtain consistent feature representations, we design dual encoding branches based on geometry and semantics to respectively map raw keypoints and image inputs into the common multi-scale feature space.

\subsubsection{Semantic Branch}
Given a pair of images $I_{\text{ref}},I_{\text{tgt}} \in \mathbb{R}^{3 \times H_0 \times W_0}$, the semantic branch extracts hierarchical semantic features. We use ResNet-18~\cite{he2016deep} as the backbone to construct multi-scale representations at scales $sl \in \{\tfrac{1}{8},\tfrac{1}{16}\}$:
\begin{equation}
    F_\text{ref,s}^{sl},F_\text{tgt,s}^{sl} = \Phi_{\text{ResNet}}(I_{\text{ref}},I_{\text{tgt}}).
    \label{eq:sem}
\end{equation}


\subsubsection{Geometric Branch}

\begin{figure}[htpb!] 
	\centering  
	\includegraphics[width=\linewidth]{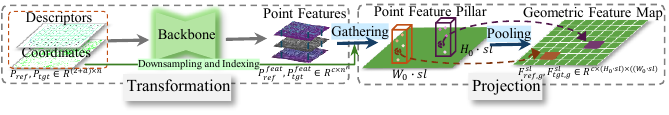} 
    \vspace{-0.6cm}
	\caption{The details of the neural point transformer module.}
	\label{fig:npt}
\end{figure}

As shown in Fig. \ref{fig:npt}, to handle sparse keypoints, we design a neural point transformer module that adopts ``Transformation-then-Projection" strategy to generate multi-scale geometric representations. 
It first encodes keypoint sets $P_\text{ref},P_\text{tgt}\in \mathbb{R}^{(2+d) \times n}$ (containing spatial coordinates $(x_i,y_i)$ and descriptors of $n$ points) into latent point feature representations via PointNeXt~\cite{qian2022pointnext}:
 \begin{equation}
 P_{\text{ref}}^{\text{feat}}, P_{\text{tgt}}^{\text{feat}} = \Phi_{\text{PointNeXt}}(P_{\text{ref}}, P_{\text{tgt}}),
 \end{equation}
where $P_{\text{ref}}^{\text{feat}}, P_{\text{tgt}}^{\text{feat}}\in \mathbb{R}^{ c\times n^{'}}$ represent the $c$-dimensional latent features of $n'$ downsampled points.
To align with semantic feature maps, we assume the geometric feature maps $F_\text{ref,g}^{sl}, F_\text{tgt,g}^{sl}$ have the same spatial dimensions as $F_\text{ref,s}^{sl}, F_\text{tgt,s}^{sl}$ (\textit{i.e.}, $\small{H_0\cdot sl \times W_0\cdot sl}$). In this hypothesis, every element in the geometric maps covers a grid cell of $\tfrac{1}{sl} \times \tfrac{1}{sl}$ pixels in the original images. Therefore, we then project $P_{\text{ref}}^{\text{feat}}, P_{\text{tgt}}^{\text{feat}}$ onto zero-initialized grids $\mathbf{V}\in \mathbb{R}^{c \times H_0\cdot sl \times W_0\cdot sl}$, in which the keypoint coordinates can be requantized as $\tilde{x}_i = \lfloor x_i \cdot sl \rfloor$ and $\tilde{y}_i = \lfloor y_i \cdot sl \rfloor$. $\lfloor \rfloor$ denotes the floor function. For grid cells containing multiple keypoints, we apply a max-pooling operation to retain the most salient features as:
 \begin{equation}
 \mathbf{V}[:, \tilde{y}, \tilde{x}] = \max_{k \in \mathcal{C}(\tilde{x}, \tilde{y})} \mathbf{f}_k,
 \label{eq3}
 \end{equation}
 where $\mathcal{C}(\tilde{x}, \tilde{y})$ is the keypoint index set within a specific cell and $\mathbf{f}_k$ is the corresponding feature vector in $P_{\text{ref}}^{\text{feat}},P_{\text{tgt}}^{\text{feat}}$. 
According to different scales, the reprojected grids following Eq.~\ref{eq3} constitute the desired geometric feature maps $F_{\text{ref,g}}^{sl}$,$F_{\text{tgt,g}}^{sl}$.

\subsection{Multimodal Feature Fusion}

After feature alignment, we aim to integrate the strengths of different representations: semantic features provide comprehensive global context but are prone to error propagation, while point features offer superior noise resistance despite their non-uniform distribution. The core challenge in achieving a balanced representation lies in harmonizing these modalities without introducing undesirable cross-modal conflicts. We contend that the solution rests upon maintaining a dynamic balance between modal independence and complementarity. To this end, we propose an adaptive mixture of experts module with a latent-space modality robustifier strategy to ensure robust feature aggregation.

\subsubsection{Adaptive Mixture of Experts}

The mixture of experts~\cite{yang2024multi} paradigm is typically applied in multi-task learning to balance the relative significance of competitive objectives. In this work, we adapt this framework for multimodal fusion to adaptively weigh the contributions of heterogeneous modal features. Our AMoE consists of three residual-based experts—semantic ($E_\text{s}$), geometric ($E_\text{g}$), and heterogeneous ($E_\text{h}$)—governed by a linear gated router $R$.
Given a pair of feature maps $(F_\text{s}^{sl}, F_\text{g}^{sl})$ at scale $sl$, the router generates a weight vector $\mathbf{w}^{sl} \in \mathbb{R}^3$:
\begin{equation}
    \mathbf{w}^{sl} = [w_\text{s}^{sl}, w_\text{g}^{sl}, w_\text{h}^{sl}] = \text{Softmax}\left( R^{sl}\left(F_{\text{s}}^{sl}, F_{\text{g}}^{sl},F_{\text{s}}^{sl}\oplus F_{\text{g}}^{sl} \right) \right),
\end{equation}
where $w_\text{s}^{sl}+w_\text{g}^{sl}+w_\text{h}^{sl} = 1$, and $\oplus$ denotes the concatenation operation. Simultaneously, the experts generate corresponding feature maps as:
\begin{equation}
\mathbf{H}_\text{s}^{sl} = E_\text{s}^{sl}(F_\text{s}^{sl}), \quad \mathbf{H}_\text{g}^{sl} = E_\text{g}^{sl}(F_\text{g}^{sl}), \quad \mathbf{H}_\text{h}^{sl} = E_\text{h}^{sl}(F_\text{s}^{sl}\oplus F_\text{g}^{sl}).
\end{equation}
The final fused feature maps $\mathbf{F}_{\text{f}}^{sl}$ is the sum of weighted aggregation as follows:
\begin{equation}
\mathbf{F}_{\text{f}}^{sl} = w_\text{s}^{sl} \mathbf{H}_\text{s}^{sl} + w_\text{g}^{sl} \mathbf{H}_\text{g}^{sl} + w_\text{h}^{sl} \mathbf{H}_\text{h}^{sl}.
\end{equation}

\subsubsection{Modality Robustifier}
Although the proposed AMoE effectively balances multimodal importance, training on complete, paired multimodal data often leads to over-reliance on modal coupling, neglecting unimodal independence. To prevent performance degradation when one modality becomes unreliable, we propose a latent-space modality robustifier strategy.

Rather than applying augmentation at the input level, which risks disrupting the delicate spatial correspondence between pixels and keypoints, we perform regularization in the latent feature space. It could enhance the robustness in challenging scenes while preserving the capability of multimodal alignment.

To this end, we introduce an additional training phase, \textit{i.e.}, freezing the alignment branches while applying random modal dropout or stochastic Gaussian noise across multi-scale features for all expert branches $r \in \{s, g, h\}$. Denoting $\epsilon \sim \mathcal{N}(0, \sigma^2 \mathbf{I})$, the perturbation process is defined as:
\begin{equation}
    \mathbf{H}_r^{sl} = 
    \begin{cases} 
    \mathbf{H}_r^{sl} & \text{if } 1-p_{\text{drop}} \\ 
    \mathbf{0} & \text{if } p_{\text{drop}} 
    \end{cases} \quad \text{or}\quad 
    \mathbf{H}_r^{sl} = 
    \begin{cases} 
    \mathbf{H}_r^{sl} & \text{if } 1-p_{\text{noise}} \\ 
    \mathbf{H}_r^{sl}+ \boldsymbol{\epsilon} & \text{if } p_{\text{noise}} 
    \end{cases}.
\end{equation}
Empirically, $p_{\text{drop}}$ and $p_{\text{noise}}$ are set to 0.25, which forces the model to construct robust representations, especially under partial modal failure or noise.

\subsection{Global-to-local Warp}
With the fused feature representations, the objective shifts to regressing precise parametric warps for content alignment and shape preservation. Similar to StabStitch++, we adopt a hierarchical global-to-local paradigm to sequentially predict homography and TPS transformations.

\subsubsection{Homography Regression}
In this part, we apply the homography decomposition algorithm~\cite{nie2025stabstitch++} to the estimated homography, which is decoupled into $\mathcal{H}_{ref}$ and $\mathcal{H}_{tgt}$. They project the input images onto a virtual middle plane, providing the initial control point offsets for subsequent TPS transformation.

\subsubsection{FFD-based TPS Regression}
With global homography, we predict residual control point offsets for local TPS warps. However, there is a memory bottleneck for TPS transformation in high-resolution. Denoting the projection matrix as $\mathbf{T}$ and output coordinate meshgrid as $\mathbf{M}$, the transformed flows can be approximately expressed as $\mathbf{X}_{\text{flow}}\sim\mathbf{T}\times \mathbf{M}$, in which a massive intermediate caching matrix could be generated.

To avoid it, we reformulate the projection as a compress-then-restore process.
We first compress the high-resolution meshgrid $\mathbf{M}$ into a low-resolution $\mathbf{M}^{'}$, with its dimensions empirically set to twice the TPS control point resolution, guaranteeing $\dim(\mathbf{M}^{'}) \ll \dim(\mathbf{M})$. We then obtain the low-resolution transformed flows ($\mathbf{X}^{'}_{\text{flow}}\sim\mathbf{T}\times \mathbf{M}^{'}$), and employ FFD to interpolate the sparse flows back to the original output scale. FFD uses the local support property of B-splines, which ensures that each pixel's deformation can be restored from only a $4 \times 4$ local lattice neighborhood as:
\begin{equation}
\mathbf{\hat{X}}_{\text{flow}}(u,v) = \sum_{\alpha=0}^{3} \sum_{\beta=0}^{3} N_{\alpha,3}(u) N_{\beta,3}(v) \mathbf{X}^{'}_{\text{flow}}(u'+\alpha-1, v'+\beta-1).
\end{equation}
$(u', v')$ represents the coordinate of an arbitrary point in $\mathbf{M}^{'}$, while $(u, v)$ denotes the pixel coordinate of any point within a specific cell of $\mathbf{M}$ mapped from the location $(u', v')$ in $\mathbf{M}^{’}$.
$N_{\cdot,3}$ denotes the cubic B-spline basis functions, which ensure the interpolated flow is continuous and smooth, and more importantly, $\mathbf{\hat{X}}_{\text{flow}}$ approximates $\mathbf{X}_{\text{flow}}$. Refer to the appendix for more details.

\subsection{Objective Function}
The objective function $\mathcal{L}$ of UniStitch comprises three terms: content alignment, shape preservation, and expert regularization within AMoE, which is defined as:
\begin{equation}
    \mathcal{L} = \mathcal{L}_{align} + w_{s}\mathcal{L}_{shape} + w_{r}\mathcal{L}_{reg.}.
    \label{eq:sdl}
\end{equation}
$\mathcal{L}_{align}$ and $\mathcal{L}_{shape}$ denote the content alignment and shape preservation losses, which follow the same formulation as StabStitch++ for a fair comparison. 

$\mathcal{L}_{reg.}$ is the expert regularization loss to ensure the most significant modal features are selected by the AMoE. We define it as:

\begin{equation}
\mathcal{L}_{reg.} = \sum_{r \in \{g,s,h\}}(w_r^{sl} - \bar{w}^{sl})^2 + \lambda_{e}\sum_{r \in \{g,s,h\}}w_r^{sl}\log(w_r^{sl}),
\end{equation}
where $\bar{w}^{sl}$ denotes the mean weight across all experts at scale $sl$. The first term imposes a variance-based penalty to ensure balanced expert utilization, while the second term, the negative entropy, promotes specialization by encouraging experts to become highly selective for specific input modalities. In our implementation, we only constrain it at the $\frac{1}{16}$ scale, \textit{i.e.}, $sl=\frac{1}{16}$.

For bravery, we put the details of $\mathcal{L}_{align}$ and $\mathcal{L}_{shape}$ in the Appendix. The hyperparameters $w_{s}$, $w_{b}$, and $\lambda_{e}$ are empirically set to 10, 0.01, and 0.1.

\section{Experiment}
\subsection{Detail and Dataset}

\subsubsection{Detail} 
We implemented UniStitch using PyTorch with a single NVIDIA RTX 5090 GPU. The training process followed a two-stage strategy.  In the first stage, we trained the model from scratch for 100 epochs without the modality robustifier. In the second stage, we continue training the model for another 100 epochs with the multimodal feature alignment frozen, while introducing the modality robustifier to enhance robustness against modality degradation. The batch size is set to 8, and the initial learning rate is set to $1 \times 10^{-4}$ with an exponential decay scheduler.

\subsubsection{Dataset and Metric} 
Similar to existing learning-based stitching solutions, we train UniStitch on the training set of UDIS-D~\cite{nie2021unsupervised}. To evaluate both in-domain performance and out-of-distribution (OOD) generalization, we test it on the testing set of UDIS-D and classic datasets collected in RopStitch~\cite{nie2026robust}, where the former includes 1,106 examples while the latter contains 147 cross-domain samples. 
As for objective metrics, we adopt mPSNR/mSSIM (as used in existing solutions~\cite{jin2025pixelstitch,nie2026robust}) to measure the average alignment errors in overlapping regions.

\subsection{Comparative Experiment}

To evaluate the effectiveness of integrating geometric and semantic features, we compare UniStitch against two categories of state-of-the-art methods:
\begin{itemize}
    \item \textbf{Traditional methods using geometric features:} Three representative algorithms (APAP \cite{zaragoza2013projective}, SPW \cite{liao2019single}, and LPC \cite{jia2021leveraging}) are selected, which leverage sparse geometric features to estimate warps.
    \item \textbf{Learning-based methods using semantic features:} Five learning-based stitching algorithms are selected including UDIS~\cite{nie2021unsupervised}, UDIS++~\cite{nie2023parallax}, DunHuangStitch~\cite{mei2024dunhuangstitch}, StabStitch++~\cite{nie2025stabstitch++}, and RopStitch~\cite{nie2026robust}. 
\end{itemize}

Among the learning-based categories, UDIS and DunHuangStitch only predict the global homography, whereas the others use homography-to-TPS warps. Notably, StabStitch++ is originally a video stitching framework; here, we adapt its spatial warp for comparison with the same number of control points as ours. 

For a fair visual comparison, we adopt the average fusion for all methods, highlighting ghosting or blurring artifacts caused by alignment errors.

\subsubsection{Quantitative Comparison}

Tables \ref{tab1} and \ref{tab2} provide a comprehensive evaluation of in-domain performance and OOD generalization across the UDIS and classical datasets, respectively. On the UDIS-D dataset, learning-based methods generally surpass traditional geometric algorithms, with the exception of UDIS and DunHuangStitch, due to the limitation of only homography. 
Conversely, on classic datasets, learning-based models struggle significantly in out-of-domain scenarios, with most underperforming compared to traditional methods. This suggests that semantic features lack the structural robustness inherent in geometric keypoints when transitioning to unseen environments. Even RopStitch, the latest SoTA using extra frozen semantic priors (from other large-scale datasets~\cite{deng2009imagenet}), fails to bridge this gap.
Unlike existing approaches, UniStitch explicitly combines the cross-scene structural awareness of geometric features with the contextual robustness of semantic features, outperforming existing approaches substantially in both in-domain and OOD settings.

\begin{table}[h!]
	\centering
	\caption{Quantitative comparison on the UDIS-D dataset. }
	\label{tab1}
    \scalebox{0.95}{
	\begin{tabular}{cccccc|ccccc}
		\hline
		 & \multirow{2}{*}{\textbf{Method}}  & \multicolumn{4}{c}{\textbf{mPSNR$\uparrow$}} &  \multicolumn{4}{c}{\textbf{mSSIM$\uparrow$}}\\
		\cline{3-10} &  & Easy & Moderate & Hard & Average & Easy & Moderate & Hard & Average \\
		\hline
		1 & APAP \cite{zaragoza2013projective} &26.77 & 22.88 & 18.75 & 22.39   &  0.868 & 0.770 & 0.587 & 0.726 \\
		2 &   SPW \cite{liao2019single} & 25.82 & 21.49 & 15.85 & 20.52 &  0.844 & 0.693 & 0.434 & 0.634 \\
		3 & LPC \cite{jia2021leveraging} & 25.01 & 21.27 & 17.34 & 20.82 & 0.815 & 0.673  &0.485 & 0.640 \\
		\hline
		4 & UDIS \cite{nie2021unsupervised} &23.53 & 19.73 & 17.42 & 19.94 & 0.761 & 0.545 & 0.376 & 0.542 \\
		5 & UDIS++ \cite{nie2023parallax}  &  27.58 & 23.75 & 20.04 & 23.41 &  0.880 & 0.792 & 0.632 & 0.755\\
		6 & DunHuangStitch \cite{mei2024dunhuangstitch} & 27.19 & 23.05 & 19.10 & 22.61 & 0.875 & 0.767 & 0.564 & 0.718 \\
		7 & StabStitch++ \cite{nie2025stabstitch++}  & 29.92 & 24.93  & 20.46 & 24.63 &  0.927 & 0.845  & 0.664 & 0.797 \\
        8 & RopStitch  \cite{nie2026robust} & 29.93 & 24.96 & 20.60 & 24.70 & 0.926 & 0.845 & 0.672 & 0.800\\
        9 & Ours & \textbf{30.34} & \textbf{25.37} & \textbf{20.90} & \textbf{25.07} & \textbf{0.932} & \textbf{0.857} & \textbf{0.691} & \textbf{0.813}\\
		\hline
		
	\end{tabular}
    }
\end{table}

\begin{table}[h!]
	\centering
	\caption{Quantitative comparison on classical datasets. $*$ denotes post processing with the iterative adaptation~\cite{nie2023parallax}. }
	\label{tab2}
     \scalebox{0.95}{
	\begin{tabular}{cccccc|cccc}
		\hline
		 & \multirow{2}{*}{\textbf{Method}}  & \multicolumn{4}{c}{\textbf{mPSNR$\uparrow$}} &  \multicolumn{4}{c}{\textbf{mSSIM$\uparrow$}}\\
		\cline{3-10} &  &  Easy & Moderate & Hard & Average & Easy & Moderate & Hard & Average \\
		\hline
        1 & APAP \cite{zaragoza2013projective} & 24.33 & 19.48 & 14.47 & 18.92 & 0.818 & 0.687 & 0.442 & 0.628 \\
        2 & SPW\cite{liao2019single} & 23.83 & 19.16 & 14.66 & 18.75 & 0.793 &  0.647 & 0.445 &  0.610 \\   
        3 & LPC \cite{jia2021leveraging} & 23.19 & 18.67 & 13.90 & 18.11 & 0.761 & 0.600 & 0.412 & 0.573 \\
        \hline
        4 & UDIS \cite{nie2021unsupervised} & 20.58 & 16.27 & 13.37 & 16.40 & 0.660 & 0.471 & 0.271& 0.454 \\
	5 & UDIS++ \cite{nie2023parallax}  &  21.93 &  17.53 & 13.83 & 17.36 & 0.709  &  0.526 & 0.325 &  0.500 \\
        6 & DunHuangStitch \cite{mei2024dunhuangstitch} & 22.02 & 17.44 & 13.65 & 17.29 & 0.712 & 0.534 & 0.319 & 0.501 \\
        7 & StabStitch++  \cite{nie2025stabstitch++} & 23.01  & 18.08  & 13.98 & 17.91 &  0.751 & 0.577  & 0.339 &  0.534 \\
        8 & RopStitch \cite{nie2026robust} & 23.40 & 18.54 & 14.67 & 18.44 & 0.772  & 0.607 & 0.387 & 0.568  \\
        9 & RopStitch$^{*}$  \cite{nie2026robust}  &25.40 & 19.79 & 15.48 & 19.74 & 0.839 & 0.691 & 0.465 & 0.645 \\
        10 & Ours & 23.92 & 18.87 & 14.93 & 18.80 & 0.800 & 0.643 & 0.410 & 0.596 \\
        11 & Ours$^{*}$ & \textbf{26.70} & \textbf{20.81} & \textbf{16.11} & \textbf{20.68} & \textbf{0.874} & \textbf{0.737} & \textbf{0.522} & \textbf{0.692} \\
		\hline
	\end{tabular}
    }
\end{table}

\subsubsection{Qualitative Comparison}

\begin{figure}[h!] 
	\centering  
	\includegraphics[width=\linewidth]{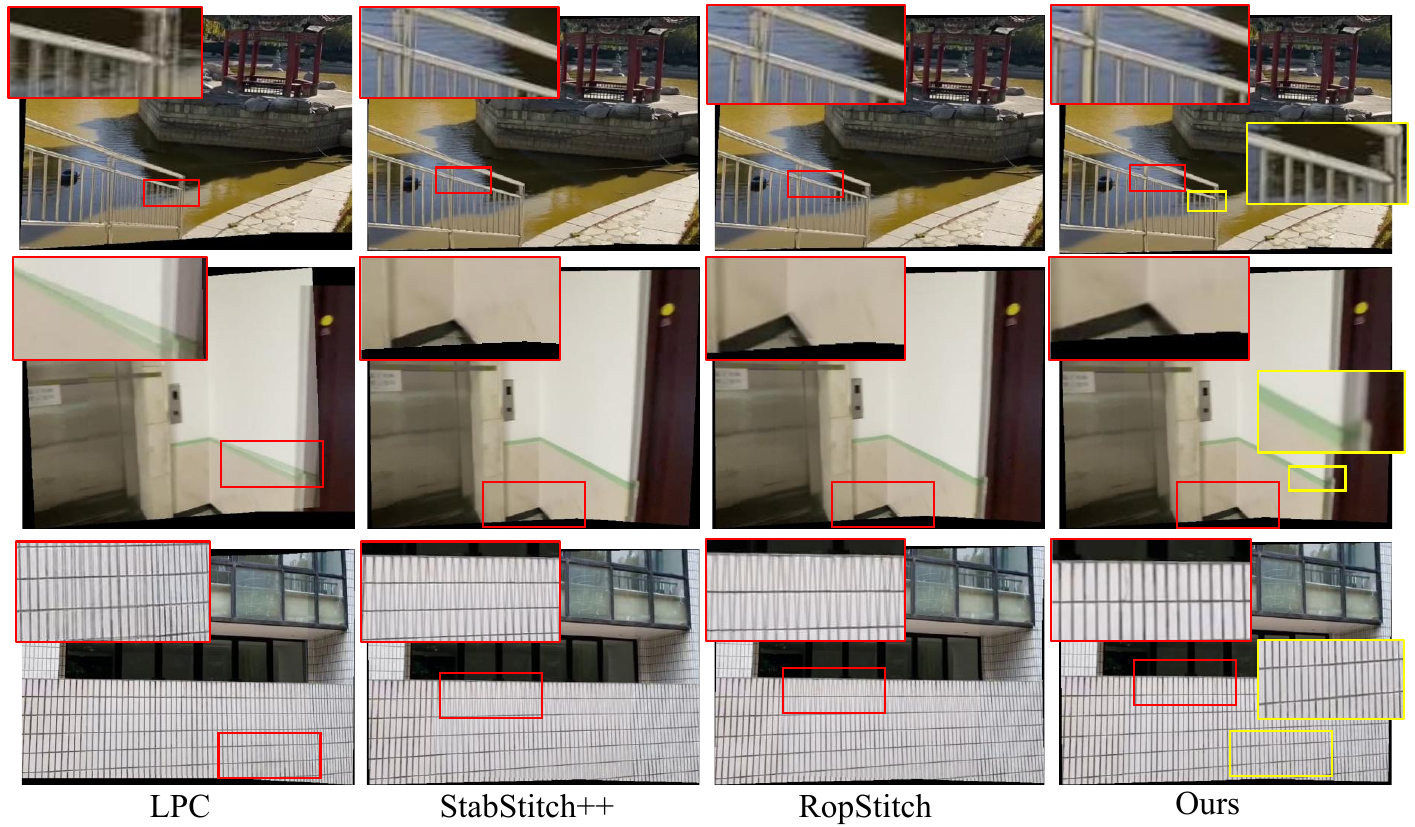} 
	\caption{Qualitative comparison on the UDIS-D dataset.}
	\label{fig:comp1}
\end{figure}

\begin{figure}[h!] 
	\centering  
	\includegraphics[width=\linewidth]{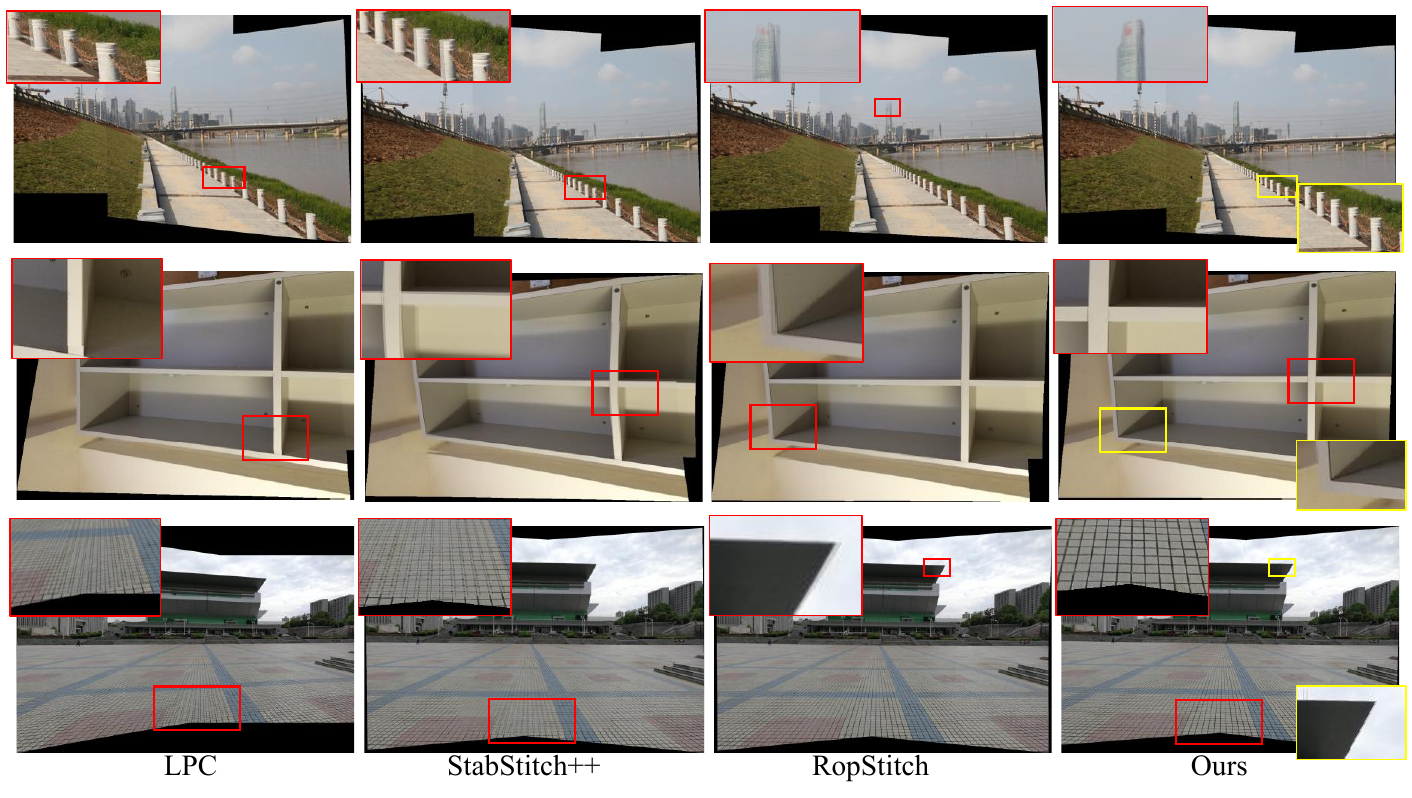} 
	\caption{Qualitative comparison on the classical dataset.}
	\label{fig:comp2}
\end{figure}

Figs. \ref{fig:comp1} and \ref{fig:comp2} present the stitching results of various algorithms on the UDIS dataset and classic datasets. It can be observed that whether dealing with the in-domain stitching of the UDIS-D dataset or the OOD generalization on classic datasets, solely relying on either semantic or geometric representations fails to achieve satisfactory performance.

Benefiting from the advantages of geometric and semantic features, UniStitch achieves superior alignment accuracy and visual coherence in these challenging scenarios. As highlighted by the red boxes, our method effectively eliminates ghosting and misalignments in complex structures such as floor tiles, railings, and distant buildings, where a single modality feature often struggles.

\subsection{Ablation Study}

\subsubsection{Impact of Network Component}
As shown in Table \ref{tab:ablation_simple}, we analyze the contribution of each core component. The point-based geometric representation and the image-based semantic representation exhibit distinct advantages: geometric features excel in OOD generalization, while semantic features perform better in in-domain scenarios. By integrating these modalities via the AMoE, we achieve significant mutual complementarity, enhancing performance across both datasets. Furthermore, the MR strategy further sharpens the model’s ability to discriminate the contributions from different modalities, highlighting the vast potential of multimodal fusion in image stitching.

\begin{table}[h]
    \centering
    \caption{Ablation study of network component.}
    \label{tab:ablation_simple}
    \begin{tabular}{ccccccc}
    \hline
    &\multicolumn{2}{c}{\textbf{Multimodal Alignment \quad}}  & \multicolumn{2}{c}{\textbf{Multimodal Fusion}} & \multirow{2}{*}{\textbf{UDIS-D}} & \multirow{2}{*}{\textbf{Classic}} \\
    \cline{1-5}
    &\textbf{Point} & \textbf{Image} & \textbf{AMoE} & \textbf{MR} & & \\
    \hline
     1&\checkmark & & & & 24.22/0.784 & 18.02/0.553 \\
     2& & \checkmark& & & 24.63/0.797 & 17.91/0.534 \\
     3&\checkmark& \checkmark& \checkmark& &24.94/0.808 & 18.56/0.581 \\
     4&\checkmark& \checkmark&\checkmark & \checkmark& \textbf{25.07}/\textbf{0.813} & \textbf{18.80}/\textbf{0.596} \\
    \hline
    \end{tabular}
\end{table}


\subsubsection{Impact of Geometric Backbones}
In Table \ref{tab:backbone_ablation}, we evaluate the impact of different point cloud encoders in a point-only set. While PointNet shows competitive in-domain results, its OOD performance is inferior to PointNet++ and PointNeXt, underscoring the superior discriminative power of multi-scale feature representations. PointTransformer v3 incorporates sophisticated scanning strategies, but its serialization of point sets may sacrifice the local positional precision required for fine-grained alignment, resulting in lower performance than other backbone networks.

\begin{table}[h]
    \centering
    \caption{Ablation study of different backbones for point feature extraction in the point-only configuration.}
    \label{tab:backbone_ablation}
    \begin{tabular}{@{}lccc@{}}
    \hline
    &\textbf{Backbone} & \textbf{UDIS-D} & \textbf{Classic} \\ 
    \hline
    1&PointNet \cite{qi2017pointnet} & \textbf{24.47}/\textbf{0.793} &17.89/0.542 \\
    2&PointNet++ \cite{qi2017pointnet++} & 24.27/0.786 & 18.00/0.551 \\
    3&\hspace{0.5em}PointTransformer v3 \cite{wu2024point} & 23.96/0.778 &  17.90/0.546 \\
    4&PointNeXt \cite{qian2022pointnext} (ours) & 24.22/0.784 & \textbf{18.02}/\textbf{0.553} \\
    \hline
    \end{tabular}
    \vspace{-0.1cm}
\end{table}

\subsubsection{Impact of Point Configuration}
The impact of point configuration is summarized in Table \ref{tab:point_order_ablation}. The model exhibits order-invariance, which is a key factor in the robustness of geometric features, analogous to the scale-invariance found in SIFT-like descriptors. Furthermore, our results emphasize point quality over quantity. In traditional pipelines, excessive outliers degrade alignment, often requiring RANSAC~\cite{fischler1981random} for pruning. Our experiments also follow this phenomenon: matched keypoint pairs yield significantly better results than raw keypoints, demonstrating that reorganizing the keypoints as correspondences is also vital for the learning-based alignment process.

\begin{table}[h]
    \centering
    \vspace{-0.2cm}
    \caption{Ablation study of different point configurations, including the point order and point quality.}
    \label{tab:point_order_ablation}
    \begin{tabular}{@{}lccc@{}}
    \hline
    &\textbf{Point Configuration} & \textbf{UDIS-D} & \textbf{Classic} \\
    \hline
    &\textbf{Training with Raw Points} & & \\
    1&-- Testing in Fixed Order & 24.78/0.803 & 17.90/0.549 \\
    2&\hspace{1.2em}-- Testing in Random Order & 24.76/0.802 & 17.90/0.549 \\
    \hline
    &\textbf{Training with Matched Points} & & \\
    3&-- Testing in Fixed Order  & \textbf{24.91}/\textbf{0.808} & \textbf{18.48}/\textbf{0.577} \\
    4&\hspace{1.2em}-- Testing in Random Order & \textbf{24.91}/\textbf{0.808} & \textbf{18.48}/0.576 \\
    \hline
    \end{tabular}
    \vspace{-0.5cm}
\end{table}

\subsubsection{Impact of Fusion Strategies}
Table \ref{tab:fusion_ablation} investigates various multimodal feature fusion methods. Simple operations like Addition or Concatenation struggle to reconcile the disparate nature of geometric and semantic features. While CSPFusion, a conventional convolutional fusion mode in YOLOv11~\cite{khanam2024yolov11}, integrates multimodal features more effectively, its OOD performance remains suboptimal. Unlike CSPFusion’s monolithic design, the proposed AMoE architecture enables differentiated processing of modalities by adaptively weighting expert networks based on the reliability of each modality. This capability is further enhanced by the MR strategy, resulting in substantial gains in OOD generalization.

\begin{table}[h]
    \centering
    \vspace{-0.1cm}
    \caption{Ablation Study of multimodal feature fusion methods. Refer to Table~\ref{tab:ablation_simple} for the results obtained with single-modality features.}
    \label{tab:fusion_ablation}
    \begin{tabular}{@{}lccc@{}}
    \hline
    &\textbf{Fusion Strategy} & \textbf{UDIS-D} & \textbf{Classic} \\ 
    \hline
    1&Add & 24.42/0.790 &  17.22/0.510 \\
    2&Concat & 24.61/0.795 & 17.74/0.538 \\
    3&CSPFusion & 24.90/0.808 & 18.52/0.576 \\
    \hline
    4&AMoE & 24.94/0.808 & 18.56/0.581 \\
    5&\hspace{0.5em}AMoE + MR (ours) & \hspace{0.5em}\textbf{25.07}/\textbf{0.813} & \hspace{0.5em}\textbf{18.80}/\textbf{0.596} \\
    \hline
    \end{tabular}
    \vspace{-0.4cm}
\end{table}

\subsubsection{Impact of FFD-based TPS}
Table \ref{tab:efficiency_comparison} compares the performance of vanilla TPS against our FFD-based TPS strategy. Results are averaged over 10 rounds, with peak GPU memory recorded. Compared to vanilla TPS, FFD-based TPS achieves substantial improvements in both inference speed and VRAM efficiency without compromising alignment accuracy (mPSNR/mSSIM). Moreover, the empirical results demonstrate that FFD-based TPS exhibits excellent resolution scalability: by decoupling deformation complexity from pixel resolution via FFD, UniStitch can effectively scale to high-resolution stitching tasks where vanilla TPS fails due to out-of-memory (OOM) errors.

\begin{table}[h!]
    \centering
    \vspace{-0.1cm}
    \caption{Ablation Study of FFD-based TPS warping strategy. Compared with vanilla TPS,  our method offers higher efficiency and lower memory usage without sacrificing alignment accuracy. }
    \label{tab:efficiency_comparison}
    \begin{tabular}{@{}lccccc@{}}
    \hline
    \textbf{Resolution} & \textbf{Warping} & \textbf{Time (s)$\downarrow$} & \textbf{Peak GPU(GB)} & \textbf{mPSNR$\uparrow$} & \textbf{mSSIM$\uparrow$} \\ 
    \hline
    \multirow{2}{*}{$566 \times 800$} & Vanilla TPS & 0.141 & 4.81 & 16.85 & 0.776 \\
     & FFD-based TPS & 0.132 & 2.76 & 16.86 & 0.776 \\
    \hline
    \multirow{2}{*}{$1329 \times 2000$} & Vanilla TPS & 0.202 & 13.99 & 17.14 & 0.475 \\
     & FFD-based TPS & 0.177 & 8.57 & 17.12 & 0.474 \\
    \hline
    \multirow{2}{*}{$2448 \times 3264$} & Vanilla TPS & - & OOM & - & - \\
     & FFD-based TPS & 0.386 & 22.62 & 22.71 & 0.627 \\
    \hline
    \end{tabular}
    \vspace{-0.5cm}
\end{table}

\subsection{Universality on Geometric Features}

To demonstrate the universality of UniStitch, we evaluate the benefits of different geometric features, including SIFT, SURF, ORB, and SuperPoint. 
As revealed in Table~\ref{tab:point_order_ablation}, matched keypoints bring more improvements than row keypoints. We apply keypoint matching to each geometric feature. Concretely, we use a traditional matching algorithm (\textit{i.e.}, KNN) for traditional geometric features and learning-based matching methods (\textit{i.e.}, SuperGlue and LightGlue) for the learning-based geometric features.

\begin{figure}[h!] 
	\centering  
	\includegraphics[width=\linewidth]{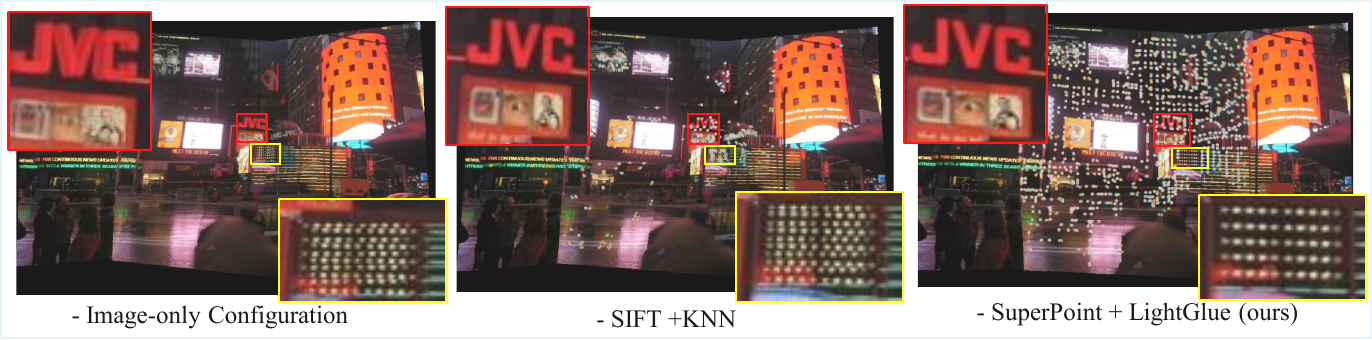} 
	\caption{Ablation study of different geometric features. We draw the matched points on the stitched results to illustrate the keypoint quality. Several local regions are zoomed in to demonstrate the benefit of extra geometric features using our method.}
	\label{fig:compare_point}
    \vspace{-0.3cm}
\end{figure}

As summarized in Table~\ref{tab:feature_ablation}, consistent performance gains across both in-domain (UDIS-D) and out-of-distribution (Classic) datasets confirm that UniStitch robustly leverages diverse geometric priors to complement semantic information. Compared with traditional features, learning-based keypoints demonstrate better results because they can establish more reliable correspondences, especially in challenging scenes (Fig. \ref{fig:compare_point} shows a corresponding example). Besides, from the last two experiments in Table~\ref{tab:feature_ablation}, we prove that feature descriptors can be crucial for disambiguating point-based representations. 

\begin{table}[h]
    \centering
    \vspace{-0.2cm}
    \caption{The impact of various geometric features. $^\ddagger$ denotes experiments that use only keypoint coordinates, without their corresponding descriptors.}
    \label{tab:feature_ablation}
    \begin{tabular}{c@{}l@{}lcc}
    \hline
    &\textbf{Feature Configuration} & \textbf{UDIS-D} & \textbf{Classic} \\ 
    \hline
    &\textbf{Baseline } & & \\
    1&\hspace{0.5em}-- Image-only Configuration & 24.63/0.797 & 17.91/0.534 \\
    \hline
    &\textbf{Tranditional } & & \\
    2&\hspace{0.5em}-- SIFT \cite{lowe2004distinctive} + KNN & 24.84/0.805 & 18.17/0.560 \\ 
    3&\hspace{0.5em}-- SURF \cite{bay2006surf} + KNN& 24.81/0.804 & 18.14/0.561 \\ 
    4&\hspace{0.5em}-- ORB \cite{rublee2011orb} + KNN& 24.85/0.806 & 18.22/0.560 \\ 
    \hline
    &\textbf{Learning-based} & & \\
    5&\hspace{0.5em}-- SuperPoint~\cite{detone2018superpoint}+SuperGlue \cite{sarlin2020superglue} & 24.85/0.806 & 18.19/0.568 \\ 
    6&\hspace{0.5em}-- SuperPoint$^\ddagger$~\cite{detone2018superpoint}+LightGlue \cite{lindenberger2023lightglue} &  24.87/0.807 & 18.43/0.573 \\ 
    7&\hspace{0.5em}-- SuperPoint~\cite{detone2018superpoint}+LightGlue \cite{lindenberger2023lightglue} (ours)\hspace{0.5em} & \textbf{24.91}/\textbf{0.808}\hspace{0.5em} & \textbf{18.48}/\textbf{0.577} \\
    \hline
    \end{tabular}
    \vspace{-0.5cm}
\end{table}

\section{Conclusion}
In this paper, we present UniStitch, a pioneering attempt to integrate geometric and semantic features into a unified image stitching model. By designing multimodal feature alignment and fusion methods, UniStitch effectively bridges the gap between the structural awareness of traditional geometric keypoints and the contextual understanding of learning-based semantic maps. Our architecture ensures that sparse geometric constraints and dense semantic priors complement each other, overcoming the limitations of single-modality approaches in texture-less or unseen environments. Extensive experiments demonstrate the superiority of UniStitch in both in-domain and out-of-distribution scenarios, revealing the promising potential of unifying multimodal features in the field of image stitching. Besides, combined with different geometric features, UniStitch achieves universal and consistent performance gains.



%
%
\bibliographystyle{splncs04}
\bibliography{main}

\appendix  
\section{Appendix}

In this document, we provide the following supplementary content:

\begin{itemize}
    \item Details of the homography decomposition. (Sec.~\ref{sec:homo})
    \item Details of the FFD and B-spline basis functions (Sec.~\ref{sec:ffd}).
    \item Details of the objective functions (Sec.~\ref{sec:loss}).
    \item Evaluation metrics (Sec.~\ref{sec:metric}).
    \item More experiments (Sec.~\ref{sec:exp}).
\end{itemize}

\section{Homography Decomposition}
\label{sec:homo}
For homography estimation, the fused feature maps ($\mathbf{F}_{\text{ref,f}}^{1/16}$ and $\mathbf{F}_{\text{tgt,f}}^{1/16}$) are first employed to construct a contextual correlation volume~\cite{nie2021depth}, which is subsequently fed into a regression network to predict the 4pt offsets~\cite{detone2016deep} of the homography matrix $\mathcal{H}$:
\begin{equation}
\mathcal{H} = \begin{pmatrix}
 h_{11} & h_{12} & h_{13}\\
 h_{21} & h_{22} & h_{23}\\
 h_{31} & h_{32} &h_{33}
\end{pmatrix} \sim \begin{pmatrix}
 \Delta u_1& \Delta v_1 \\
 \Delta u_2& \Delta v_2 \\
 \Delta u_3& \Delta v_3\\
\Delta u_4& \Delta v_4
\end{pmatrix},
\end{equation}
where $(\Delta u_i,\Delta v_i)$ denotes the vertex offsets. Using the four vertices and their motions, we can get four pairs of matched points, uniquely determining a homography transformation. Then, similar to StabStitch++~\cite{nie2025stabstitch++}, we can determine a virtual middle plane by halving all the displacements from the 4pt representation:
\begin{equation}
\mathcal{H}_{tgt} \sim \begin{pmatrix}
 \Delta u_1/2& \Delta v_1/2 \\
 \Delta u_2/2& \Delta v_2/2 \\
 \Delta u_3/2& \Delta v_3/2\\
\Delta u_4/2& \Delta v_4/2
\end{pmatrix}.
\end{equation}
$\mathcal{H}_{\text{tgt}}$ represents the projection from the target plane to the middle plane. Accordingly, the homography mapped from the reference plane to the middle can be calculated as:
\begin{equation}
\mathcal{H}_{\text{ref}} = \mathcal{H}^{-1}\mathcal{H}_{\text{tgt}}.
\end{equation}

Next, the bidirectional transformations ($\mathcal{H}_{\text{ref}},\mathcal{H}_{\text{tgt}}$) can, on one hand, provide initial positions for predicting local TPS control points, and on the other hand, warp the feature maps ($\mathbf{F}_{\text{ref,f}}^{1/8}, \mathbf{F}_{\text{tgt,f}}^{1/8}$). We then adopt the warped features to calculate local cost volumes~\cite{sun2018pwc}, which are used to estimate residual offsets of TPS control points for two views.

\section{FFD and B-Spline Basis Functions}
\label{sec:ffd}
The FFD leverages cubic B-splines as basis functions for spatial interpolation. These functions ensure that the resulting displacement field is continuous, providing high-order smoothness for image stitching. The basis functions $N_{\alpha,3}(x)$ ($\alpha \in \{0, 1, 2, 3\}$) can be expressed in a compact matrix form as follows:

\begin{equation}
    \begin{bmatrix}
    N_{0,3}(x) \\
    N_{1,3}(x) \\
    N_{2,3}(x) \\
    N_{3,3}(x)
    \end{bmatrix}
    = \frac{1}{6}
    \begin{bmatrix}
    1 & -3 & 3 & -1 \\
    4 & 0 & -6 & 3 \\
    1 & 3 & 3 & -3 \\
    0 & 0 & 0 & 1
    \end{bmatrix}
    \begin{bmatrix}
    1 \\
    x \\
    x^2 \\
    x^3
    \end{bmatrix},
\end{equation}
This local support property allows the deformation at any pixel to be computed as a weighted sum of the $4 \times 4$ control points surrounding it. 

\section{Objective Function}
\label{sec:loss}
\subsection{Content Alignment Constraint}

This constraint facilitates both coarse and fine-grained alignment by combining global homography-based transformations ($\mathcal{H}_{\text{ref}}$, $\mathcal{H}_{\text{tgt}}$) with local TPS warps ($\mathcal{T}_{\text{ref}}$, $\mathcal{T}_{\text{tgt}}$):

\begin{equation}
\begin{aligned}
\mathcal{L}_{align} = & \lambda_\mathcal{H}||\mathcal{W}(I_{\text{ref}}, \mathcal{H}_{\text{ref}}) - \mathcal{W}(I_{\text{tgt}}, \mathcal{H}_{\text{tgt}} )||_{1}\odot M_{\mathcal{H}} +   \\& \lambda_\mathcal{T}||\mathcal{W}(I_{\text{ref}}, \mathcal{T}_{\text{ref}}) - \mathcal{W}(I_{\text{tgt}}, \mathcal{T}_{\text{tgt}} )||_{1}\odot M_{\mathcal{T}},   \quad \text{where}\\
\end{aligned}
\end{equation}
\begin{equation}
\begin{aligned}
M_{\mathcal{H}} = &\mathcal{W}(\mathbb{I}, \mathcal{H}_{\text{ref}}) \cdot \mathcal{W}(\mathbb{I}, \mathcal{H}_{\text{tgt}} ), \quad M_{\mathcal{T}} = \mathcal{W}(\mathbb{I}, \mathcal{T}_{\text{ref}}) \cdot \mathcal{W}(\mathbb{I}, \mathcal{T}_{\text{tgt}} ),
\end{aligned}
\end{equation}
where $\mathcal{W}(\cdot,\cdot)$ denotes the warping operation. The binary masks $M_\mathcal{H}$ and $M_\mathcal{T}$ are derived by warping an all-ones matrix $\mathbb{I}$, identifying the overlapping regions to 1 and non-overlapping regions to 0. We set $\lambda_\mathcal{H}=1$ and $\lambda_\mathcal{T}=3$ to balance the significance of global and local warps.

\subsection{Shape-preserving Constraint}

The shape term includes an intra-grid constraint and an inter-grid constraint as:
\begin{equation}
	\mathcal{L}_{shape} = \mathcal{L}_{shape}^{intra} + \mathcal{L}_{shape}^{inter}. 
\end{equation}
This allows for a reasonable magnitude of image warping and ensures consistent motions of neighboring control points. Denoting the resolution of input images as $H\times W$, we define the TPS control points evenly distributed on the whole image with $(U+1)\times(V+1)$ points. These control points can be regarded as the vertices of a $U\times V$ mesh.

\textbf{Intra-grid Constraint:} 
To prevent excessive deformation of the mesh, we set a minimum length and width in both horizontal and vertical directions. This constraint restricts the generated mesh from being too small and causing dramatic scaling. Denoting $\svec{i}$ and $\svec{j}$ as the unit vectors in the horizontal and vertical directions, we can formulate this constraint as:

\begin{equation}
  \begin{matrix}
    \begin{aligned}
      \mathcal{L}_{shape}^{intra} = &\frac{1}{^{(U+1) \times V}} \sum{L_{h}^{intra}} + \frac{1}{^{U \times (V+1)}} \sum{L_{v}^{intra}}, \quad\text{where}\\
    \end{aligned}
  \end{matrix}
\end{equation}

\begin{equation}
  \begin{matrix}
    \begin{aligned}
      \mathcal{L}_{h}^{intra} = ReLU(|<\svec{e}_{h},\svec{i}>| - \alpha \frac{_W}{^V}),\\
       \mathcal{L}_{v}^{intra} = ReLU(|<\svec{e}_{v},\svec{j}>| - \alpha \frac{_H}{^U}).
    \end{aligned}
  \end{matrix}
\end{equation}
$L_{h}^{intra}$ and $L_{v}^{intra}$ represent the intra-grid constraint terms in the horizontal and vertical directions, respectively. $\svec{e}_{h}$ and $\svec{e}_{v}$ represent the direction vectors connected by neighboring points in the horizontal and vertical directions, respectively. $\alpha$ is set to $1/2$, and both $U$ and $V$ are set to 12.

\textbf{Inter-grid Constraint:} 
To preserve the structural integrity of straight lines and ensure that global collinearity is maintained after warping, we enforce a smoothness constraint across neighboring grids. This is achieved by penalizing the angular deviation between successive edges in the mesh, thereby encouraging neighboring cells to undergo consistent geometric transformations.

For two successive horizontal edges $\svec{e}^{\,w}_{i,j}$ and $\svec{e}^{\,w}_{i,j+1}$, the angular error $\epsilon^w_{i,j}$ is derived from their cosine similarity:
\begin{equation}
\epsilon^w_{i,j} = 1 - \frac{<\svec{e}^{\,w}_{i,j},\svec{e}^{\,w}_{i,j+1}>}{\|\svec{e}^{\,w}_{i,j}\|_2 \|\svec{e}^{\,w}_{i,j+1}\|_2}.
\end{equation}
Similarly, for vertical edges $\svec{e}^{\,h}_{i,j}$ and $\svec{e}^{\,h}_{i+1,j}$, the vertical error is defined as:
\begin{equation}
\epsilon^h_{i,j} = 1 - \frac{<\svec{e}^{\,h}_{i,j}, \svec{e}^{\,h}_{i+1,j}>}{\|\svec{e}^{\,h}_{i,j}\|_2 \|\svec{e}^{\,h}_{i+1,j}\|_2}.
\end{equation}

To provide a more holistic regularization of the mesh, we aggregate these edge-wise errors to define the distortion per grid cell (quad). The horizontal and vertical quad-level errors are formulated as:
\begin{equation}
\Delta^w_{i,j} = \epsilon^w_{i,j} + \epsilon^w_{i+1,j}, \quad\Delta^h_{i,j} = \epsilon^h_{i,j} + \epsilon^h_{i,j+1}.
\end{equation}

The final inter-grid loss is computed as the mean of these accumulated errors across all horizontal and vertical quad segments:

\begin{equation}
\mathcal{L}_{shape}^{inter} = \frac{1}{N_w} \sum_{i,j} \Delta^w_{i,j} + \frac{1}{N_h} \sum_{i,j} \Delta^h_{i,j},
\end{equation}
where $N_w$ and $N_h$ denote the total number of horizontal and vertical quad configurations, respectively. This formulation effectively penalizes localized ``bending" of the mesh, ensuring that the stitching results remain visually natural and free from geometric artifacts.

\section{Evaluation Metric} 
\label{sec:metric}
Following the methodology of StabStitch++ \cite{nie2025stabstitch++}, we use Masked Structural Similarity (mSSIM) and Masked Peak Signal-to-Noise Ratio (mPSNR) as our primary quantitative metrics. These metrics prioritize alignment accuracy by calculating values exclusively within the valid overlapping regions, thereby avoiding the bias introduced by non-overlapping background pixels.

Let $O_{ref}$ and $O_{tgt}$ denote the reference and target images warped onto the final composite canvas. The calculation for the overlapping region mask $M_{olp}$ is a binary operation where a value of 1 indicates a pixel $p$ exists in both warped images. 

\textbf{Masked SSIM:}
It measures the structural integrity between the two warped images within the overlap:

\begin{equation}
mSSIM = \frac{\sum_{p \in O} M_{olp}(p) \cdot SSIM(p)}{\sum_{p \in O} M_{olp}(p)},
\end{equation}
where $SSIM(\cdot)$ is computed locally using a standard $7 \times 7$ window.

\textbf{Masked PSNR:}
This metric evaluates the pixel-level registration fidelity, derived from the Masked Root Mean Square Error (mRMSE):
\begin{equation}
  \begin{matrix}
    \begin{aligned}
      mPSNR =& 20 \cdot \log_{10}\left(\frac{1}{mRMSE}\right),\quad \text{where}\\
      mRMSE =& \sqrt{ \frac{\sum_{p \in O} M_{olp}(p) \cdot (O_{ref}(p) - O_{tgt}(p))^2}{\sum_{p \in O} M_{olp}(p)} },
    \end{aligned}
  \end{matrix}
\end{equation}
where $p$ represents the 2D pixel coordinates on the canvas $O$. High values of mSSIM and mPSNR indicate superior alignment performance.

\section{More Experiments}
\label{sec:exp}
\subsection{Modality Robustness Analysis of AMoE}

\begin{figure}[h!] 
	\centering  
	\includegraphics[width=\linewidth]{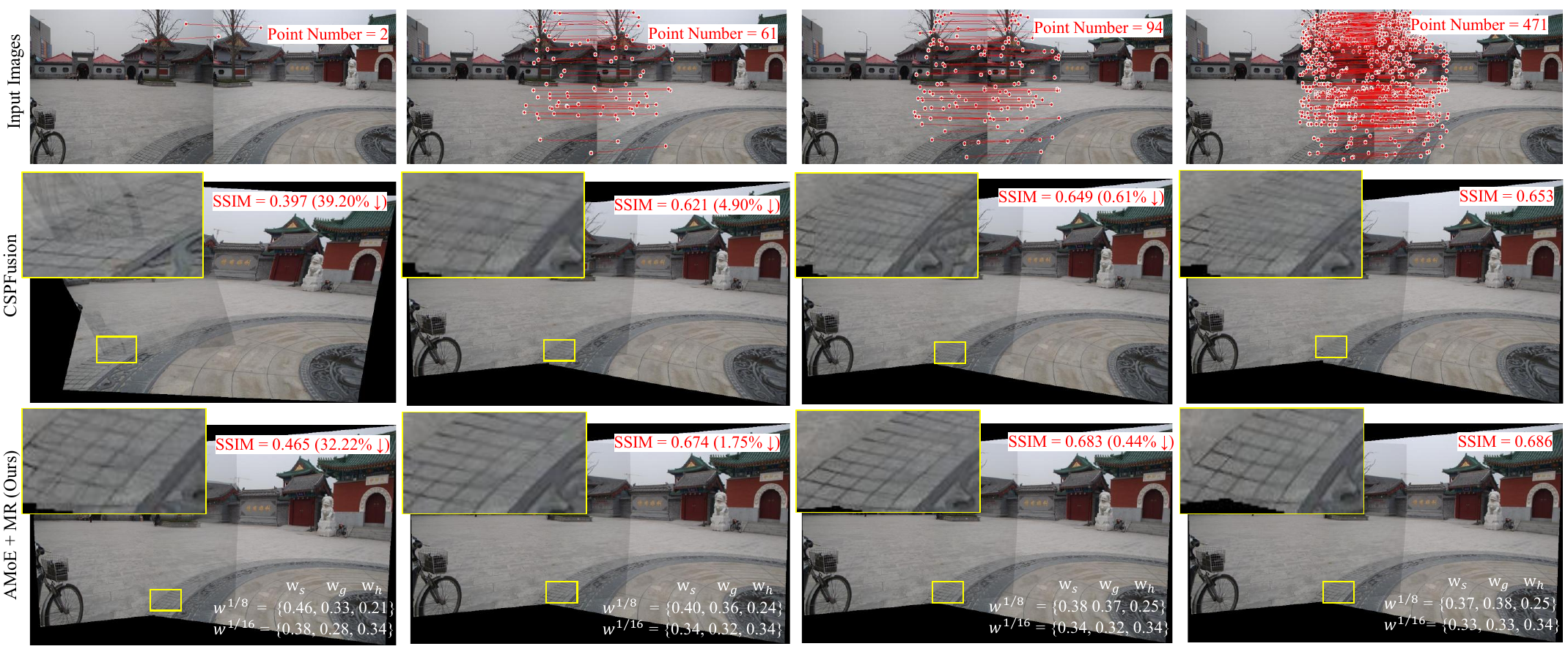} 
	\caption{Ablation study on different multimodal feature fusion methods.}
	\label{fig:fusion_methods}
\end{figure}

Fig. \ref{fig:fusion_methods} compares AMoE with conventional convolutional fusion (e.g., CSPFusion) under varying keypoint counts, where the number of points serves as a proxy for geometric reliability.  As defined in the main paper, the expert weights are $\mathbf{w}^{sl}= [w_{s}^{sl}, w_{g}^{sl}, w_{h}^{sl}]$, representing the significance of semantic, geometric, and heterogeneous experts at scale $sl$.
The results show that AMoE+MR consistently outperforms CSPFusion in stitching quality with less performance decay, which highlights its superior noise resistance. Notably, when the point count is as few as 4 (the first column of Fig.~\ref{fig:fusion_methods}), CSPFusion-based results suffer from severe distortion, while our method maintains a reasonable coarse alignment. Furthermore, the numerical changes in $\mathbf{w}^{sl}$ confirm that AMoE adaptively perceives modality reliability, assigning higher weights to the more important source.

\subsection{Impact of Point Pooling Strategies}
Table \ref{tab:pooling_ablation} compares three global aggregation strategies in the neural point transformer module. The results show that the choice of pooling (Mean, Sum, or Max) does not significantly affect the performance. It suggests that the model’s robustness is primarily derived from the intrinsic quality of the point representations rather than the specific pooling method.

\begin{table}[h!]
    \vspace{-0.1cm}
    \centering
    \caption{Ablation study of different pooling strategies.}
    \label{tab:pooling_ablation}
    \begin{tabular}{@{}lcc@{}}
    \hline
    \textbf{Pooling Strategy} & \textbf{UDIS-D} & \textbf{Classic} \\ 
    \hline
    Mean Pooling &  24.89/\textbf{0.808} & 18.45/0.576 \\
    Sum Pooling &  \textbf{24.92}/0.807 & \textbf{18.49}/\textbf{0.577} \\
    Max Pooling (ours) & 24.91/\textbf{0.808} & 18.48/\textbf{0.577} \\
    \hline
    \end{tabular}
    \vspace{-0.5cm}
\end{table}

\subsection{Impact of the Expert Regularization Loss}
Table \ref{tab:balance_loss_ablation} evaluates the impact of applying the expert regularization loss $\mathcal{L}_{reg.}$ at different feature scales. The superscript in $\mathcal{L}_{reg.}$ indicates the scale at which this constraint is applied. Results demonstrate that enforcing the constraint exclusively at the $1/16$ scale yields the optimal overall performance. This phenomenon suggests that sparse geometric point features exhibit higher representational synergy with semantic features at the $1/16$ scale, which primarily governs global registration. Since global warping is inherently robust to local point sparsification, the $1/16$ scale provides a more stable latent space for the AMoE to balance modal contributions.

\begin{table}[h!]
    \vspace{-0.1cm}
    \centering
    \caption{Ablation study of balance loss.}
    \label{tab:balance_loss_ablation}
    \begin{tabular}{@{}lcc@{}}
    \hline
    \textbf{Configuration} & \textbf{UDIS-D} & \textbf{Classic} \\ 
    \hline
    w/o $L_{reg}$ & 24.91/0.807 & 18.42/0.579 \\
    w/ $L_{reg}^{1/8}$  &  24.91/\textbf{0.808} & 18.48/0.577 \\
    w/ $L_{reg}^{1/16}$ (ours)  & \textbf{24.94}/\textbf{0.808} & \textbf{18.56}/\textbf{0.581} \\
    w/ $L_{reg}^{1/8}$ + $L_{reg}^{1/16}$ & 24.93/\textbf{0.808} & 18.39/0.568 \\
    \hline
    \end{tabular}
    \vspace{-0.5cm}
\end{table}

\subsection{Inference Time}
Table \ref{tab:inference_time} compares the inference speeds of several learning-based image stitching algorithms across varying resolutions. Aside from RopStitch, which exhibits significantly higher latency due to its multi-round retrieval process, all other evaluated methods achieve efficient inference. These results underscore the substantial potential of learning-based approaches for future real-time stitching applications.

\begin{table}[h!]
    \vspace{-0.1cm}
    \centering
    \caption{Comparisons on inference time (/s). These methods are tested with Intel(R) Core(TM) i7-14700KF CPU @ 3.40GHz and NVIDIA RTX 4090 D GPU.}
    \label{tab:inference_time}
    \begin{tabular}{lccc}
        \hline
        \textbf{Dataset} & \textbf{Cat} \cite{lee2020warping} & \textbf{Reception} \cite{jia2021leveraging} & \textbf{Construction site} \cite{zaragoza2013projective} \\
        \hline
        Resolution & 400$\times$600 & 675$\times$1200 & 1329$\times$2000 \\
        \hline
        UDIS \cite{nie2021unsupervised} & 0.0991 & 0.1339 & 0.1901 \\
        UDIS++ \cite{nie2023parallax} & 0.0548 & 0.0869 & 0.1784 \\
        DunhuangStitch \cite{mei2024dunhuangstitch} & 0.0468 & 0.1092 & 0.1271 \\
        StabStitch++ \cite{nie2025stabstitch++} & 0.0466 & 0.1262 & 0.1933 \\
        RopStitch (w/o search) \cite{nie2026robust} & 0.0389 & 0.0868 & 0.1566 \\
        RopStitch \cite{nie2026robust}  & 0.4134 & 0.9365 & 1.7433 \\
        Ours & 0.1109 & 0.1489 & 0.2010 \\
        \hline
    \end{tabular}
    \vspace{-0.5cm}
\end{table}

\subsection{More Qualitative Comparison Results}
\begin{figure}[htbp!] 
	\centering  
	\includegraphics[width=\linewidth]{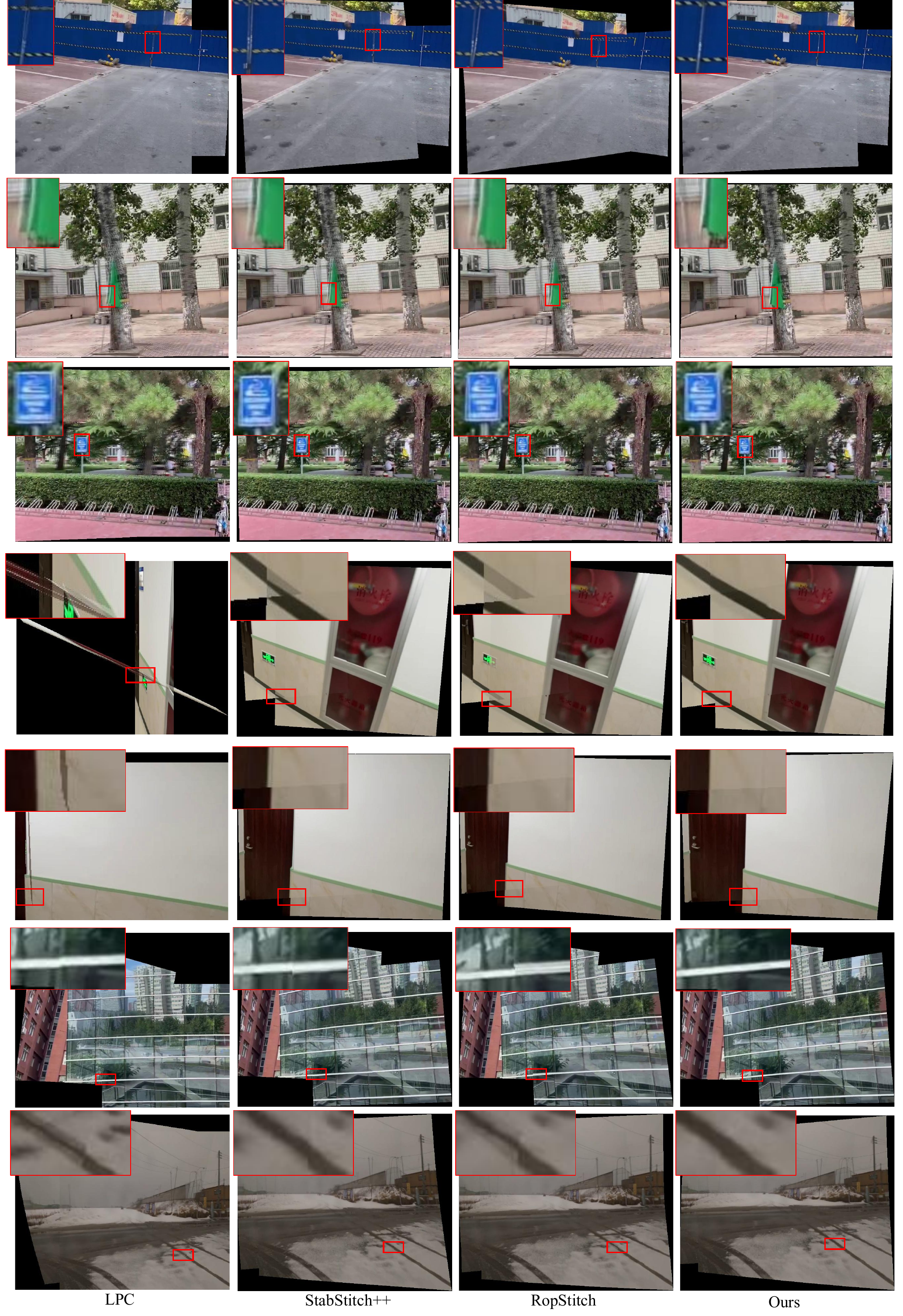} 
	\caption{More comparative results on the UDIS-D dataset.}
	\label{fig:supp_compare1}
\end{figure}

\begin{figure}[htbp!] 
	\centering  
	\includegraphics[width=\linewidth]{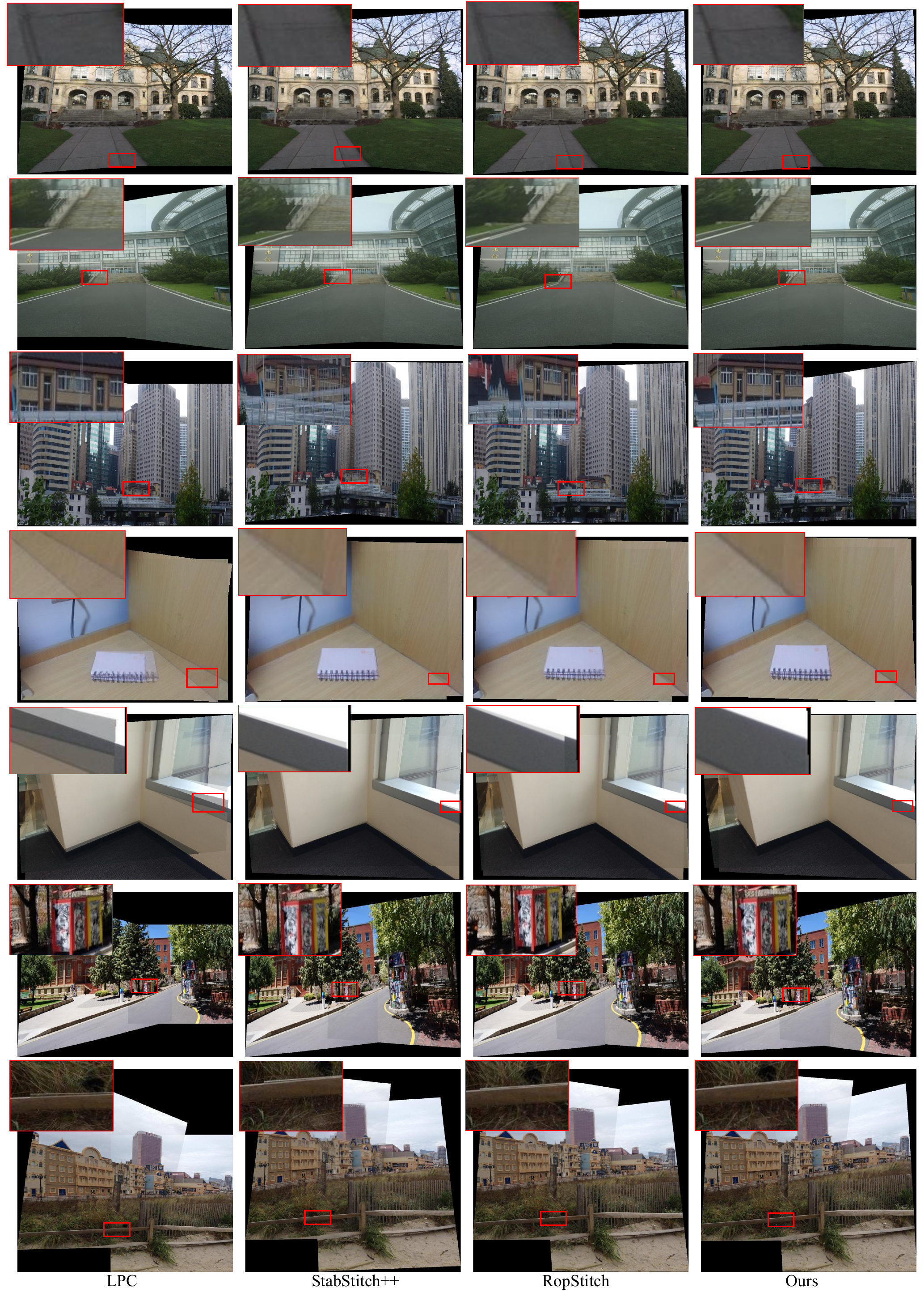} 
	\caption{More comparative results on the classical dataset.}
	\label{fig:supp_compare2}
\end{figure}

Fig. \ref{fig:supp_compare1} and \ref{fig:supp_compare2} provide additional qualitative comparisons on the UDIS-D and Classical datasets. As illustrated, our proposed algorithm consistently achieves superior alignment performance and finer detail preservation compared to existing state-of-the-art methods.

\end{document}